\documentclass[letterpaper, 10 pt, conference]{ieeeconf}  

\IEEEoverridecommandlockouts                              

\usepackage{times}
\usepackage{booktabs}
\usepackage{multicol}
\usepackage{booktabs}
\usepackage{graphics} 
\usepackage{times} 
\usepackage{amsmath} 
\usepackage{amssymb}  

\usepackage{subcaption}
\usepackage[belowskip=-1.5pt]{caption}

\usepackage[T1]{fontenc}

\usepackage{times}
\usepackage{epsfig}
\usepackage{graphicx}
\usepackage{amsmath}
\usepackage{amssymb}
\usepackage{stfloats}\usepackage{multirow}
\usepackage{cite} 
\usepackage{rotating}

\usepackage[noend]{algpseudocode}
\usepackage{algorithm}
\usepackage{xspace}
\usepackage{multirow}

\usepackage{arydshln}
\DeclareMathAlphabet{\mathpzc}{OT1}{pzc}{m}{it}

\newcommand{\mathkomma}{\quad ,}
\newcommand{\mathpunkt}{\quad .}

\usepackage{color}

\usepackage{todonotes}

\usepackage[normalem]{ulem}

\makeatletter
\newcommand\footnoteref[1]{\protected@xdef\@thefnmark{\ref{#1}}\@footnotemark}
\makeatother

\usepackage[per=slash]{siunitx}
\usepackage{comment}

\definecolor{OliveGreen}{RGB}{0,200,25}
\newcommand{\red}[1]{\textcolor{red}{#1}}

\newcommand{\darkgreen}[1]{\textcolor{OliveGreen}{#1}}

\newcommand{\ie}{i.\,e.\ }
\newcommand{\eg}{e.\,g.\ }

\newcommand{\armarIIIa}{\mbox{ARMAR-IIIa}\xspace}

\newcommand{\ackTimestormREBA}{The research leading to these results has received  funding from the European Union's Horizon 2020 research and innovation programme under grant agreement No 641100 (TimeStorm) and from the German Research Foundation (DFG: Deutsche Forschungsgemeinschaft) under Priority Program on Autonomous Learning (SPP 1527).}

\newcommand{\added}[1]{\blue{#1}}
\newcommand{\replaced}[2]{\red{\ifmmode\text{\sout{\ensuremath{#1}}}\else\sout{#1}\fi}\darkgreen{#2}}
\newcommand{\removed}[1]{\red{\ifmmode\text{\sout{\ensuremath{#1}}}\else\sout{#1}\fi}}

	\renewcommand{\added}[1]{#1}
 	\renewcommand{\replaced}[2]{#2}
	\renewcommand{\removed}[1]{}

\newcommand{\addedimage}[1]{\fcolorbox{blue}{blue}{#1}}
\newcommand{\removedfootnote}[1]{\footnote{\removed{#1}}}
\newcommand{\removedsubsection}[1]{\subsection{\texorpdfstring{\removed{#1}}{#1}}}

	\renewcommand{\addedimage}[1]{#1}
	\renewcommand{\removedfootnote}[1]{}
	\renewcommand{\removedsubsection}[1]{}
	
	\renewcommand{\removedsubsection}[1]{}

\title{\LARGE \bf
Deep Episodic Memory: Encoding, Recalling, and Predicting \\ Episodic Experiences  for Robot Action Execution
}

\author{Jonas Rothfuss$^{\ast}{^\dagger}$,
 Fabio Ferreira$^{\ast}{^\dagger}$,
  Eren Erdal Aksoy $^{\ddagger}$,
  You Zhou$^{\dagger}$ and Tamim Asfour$^{\dagger}$
\thanks{\ackTimestormREBA}
\thanks{$^{\dagger}$The authors are with the Institute for Anthropomatics and Robotics, Karlsruhe Institute of Technology, Karlsruhe, Germany.} 
\thanks{$^{\ddagger}$Eren E. Aksoy is with with the School of Information Technology, Halmstad University, Sweden}
\thanks{$^{\ast}$The first two authors contributed equally to this work.}
}

\begin{document}
\markboth{IEEE Robotics and Automation Letters. Preprint Version. Accepted Month, Year}
{FirstAuthorSurname \MakeLowercase{\textit{et al.}}: ShortTitle} 

\maketitle
\begin{abstract}

We present a novel deep neural network architecture for representing robot experiences in an episodic-like memory which facilitates encoding, recalling, and predicting action experiences.
Our proposed unsupervised deep episodic memory model 1) encodes observed actions in a latent vector space and, based on this latent encoding, 2) infers most similar episodes previously experienced, 3) reconstructs original episodes, and 4) predicts future frames in an end-to-end fashion. 
Results show that conceptually similar actions are mapped into the same region of the latent vector space.
Based on these results, we introduce an action matching and retrieval mechanism, benchmark its performance on two large-scale action datasets, 20BN-something-something and ActivityNet and evaluate its generalization capability in a real-world scenario on a humanoid robot.
\end{abstract}



\section{Introduction}
  
Humans are ingenious: We have unique abilities to predict the consequences of observed actions, remember the most relevant experiences from the past, and transfer knowledge from previous observations in order to adapt to novel situations. 
The episodic memory which encodes contextual, spatial and temporal experiences during development plays a vital role to introduce such cognitive abilities in humans. 
 
A core challenge in cognitive robotics is compact and generalizable mechanism which allow for encoding, storing and retrieving spatio-temporal patterns of visual observations. 
Such mechanisms would enable robots to build a memory system, allowing them to efficiently store gained knowledge from past experiences and both recalling and applying such knowledge in new situations. 

\begin{figure}[!t]
       \centering
       \includegraphics[width=0.63\linewidth]{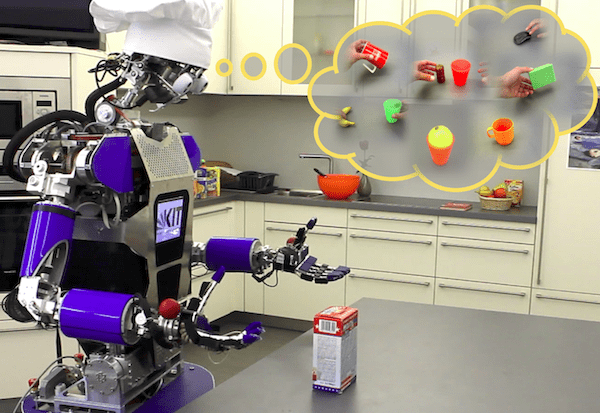}
        \caption{The \armarIIIa humanoid robot, see \kern-0.5em\cite{Asfour2006}, recalling previous visual episodes in a kitchen scene.}
\label{fig:armar_kitchen}
\end{figure}

Inspired by infants that learn by observing and memorizing what adults do in the same visual setting, we investigate in this paper how to extend cognitive abilities of robots to autonomously infer the most probable behavior and ultimately adapt it to the current scene.
Considering the situation of the humanoid robot \armarIIIa standing in front of a table with a juice carton (see Fig.~\ref{fig:armar_kitchen}) one can ask what the most suitable action is and how it would best be performed.

To achieve this goal, we introduce a novel deep neural network architecture for encoding, storing, and recalling past  action experiences in an episodic memory-like manner. 
The proposed deep network encodes observed action episodes in a lower-dimensional latent space. Such a formulation in the latent space allows robots to store visual experiences, compare them based on their conceptual similarity and retrieve the most similar episodes to the query scene or action.  
Further, the same network leads to predict and generate the next possible frames of a currently observed action.

To the best of our knowledge, this is the first study introducing that vision-based cognitive abilities concerning action representing, storing, memorizing, and predicting can be achieved in a single coherent framework.
We hypothesize that latent subsymbolic encodings that our network generates from visual observations are rich and descriptive enough to be compared with those collected from previously experienced episodes.
In this way, \armarIIIa can trace all previous observations and select the most similar episode (\eg ``pushing the juice" or ``grasping the juice") in the latent space. The robot can further generate similar behavior by adapting to new situations based on memorized action representations.

\textbf{Contribution:} 
(1) We implement a new deep network to encode action frames into a low-dimensional latent vector space.
(2) Such a vector representation is used to reconstruct the action frames in an auto-encoder manner.
(3) We show that the same latent vectors can also be employed to predict future action frames.
(4) We introduce a mechanism for matching and retrieving visual episodes and provide an evaluation of the proposed method on two action datasets.
(5) Finally, we demonstrate how this meachanism can facilitate case-based reasoning for robotic object manipulation in an unstructured real-world scenario.

\section{Related Work}
We discuss related work from two relevant perspectives: the role of episodic memories in cognitive architectures and action understanding based on deep learning approaches.

\subsection{Episodic Memory and Cognitive Architectures}
In contrast to working memory where the information is temporarily stored for a finite length of time, the long-term memory  holds the innate knowledge that enables operation of the system and facilitates learning. 
The episodic memory, considered as a part of the long-term memory, persists instances of past experiences which can be retrieved to support planning and inference  \cite{Kotseruba2016}. Hereby, the persisted experiences can be represented in manifold ways.

Reinforcement learning based architectures implement memories to store and retrieve action episodes\cite{Sun1998, Rohrer2009}.
A different approach is to persist instances of the working memory that were involved in solving a specific problem and subsequently retrieve previous solutions from the episodic memory.
Thereby, planning can be enhanced and even facilitate one-shot learning capabilities \cite{Nuxoll2007, Kuppuswamy2006, Stachowicz2012, Vernon2010, DeepART}. Predominantly, instances stored in the episodic memory are symbolic high-level representations \cite{Kuokka1991, Stachowicz2012}. 
When restricted to a specific context, symbolic representations and pre-specified perceptual instances stored in an episodic memory can indeed be a powerful approach for enhancing the reasoning capabilities of a cognitive system, as shown in Soar \cite{Nuxoll2007}. However, most of the described approaches are customized to a specific problem domain and rely on pre-defined, problem specific representations \cite{Kuokka1991, Kuppuswamy2006}. In complex real world scenarios transferring and generalizing knowledge persisted in the episodic memory is very limited when pre-defined symbolic representations are used. Accounting for nuances and fuzziness may require interpolation between concepts, demanding more flexibility than traditional declarative memory concepts. Our proposed episodic memory, on the other hand, derives subsymbolic representation of actions in a data driven manner and, hence, requires no pre-defined information.

An approach towards an episodic-like memory of video scenes, based on subsymbolic representations, uses Fisher Vectors of convolutional neural network (CNN) percepts to generate encodings of temporal video segments \cite{Doshi2015}. Although the approach is able to match conceptually related video segments, it is not possible to reconstruct perceptual information from the Fisher Vector representations. 

\subsection{Action Understanding with Deep Neural Nets}
Many deep neural network based approaches to understand human action videos combine CNNs and recurrent neural networks (RNNs) \cite{Ranzato2014, Donahue2015, Srivastava2015}.
CNNs capture spatial information in each video frame and aggregate it into a higher-level representation, which is then fed through a  Long Short-
Term Memory (LSTM) that captures temporal information throughout a sequence of frames. 
Instead of stacking an LSTM on top of a CNN, Shi et al.~\cite{Shi2015} combine the ideas of spatial weight sharing through convolution and temporal weight sharing through recurrence into a new model called convolutional LSTMs (convLSTM).

Overall, there are three main approaches with regard to deep learning based  action understanding: 1) Supervised learning on activity recognition corpora \cite{UCF101, Heilbron2015, Koppula13a}, 2) Unsupervised video frame prediction \cite{Mathieu2015, Lotter2016, Patraucean2016, Finn2016} and 3) Unsupervised prediction of visual representations \cite{Vondrick2016a}.

Aside from human action recognition, the lack of comprehensively labeled video datasets  makes supervised training challenging. Another approach towards learning to understand videos is the future frame prediction. Given a sequence of video frames, a deep neural network is trained to predict the next frame(s) in a video. To successfully predict future video frames, the network is forced to generate latent representations of the inherent structure and dynamics of videos. 

Srivastava et al. \cite{Srivastava2015} present a composite model consisting of three LSTM networks, conceptually combining a sequence-to-sequence autoencoder with future frame prediction. They show that the composite architecture outperforms both a pure autoencoder and future frame prediction model. However, since input and output space are CNN features instead of raw video frames, their model is not able to recover the video frames from the latent representation. Our approach is inspired by the composite encoder-decoder architecture but overcomes the described drawback by being able to encode raw frames in an end-to-end fashion and also reconstruct the raw frame sequence from the latent representation.

Subsequent work aims to predict the pixel changes between the current and the next frame \cite{Lotter2016, Patraucean2016, Finn2016} instead of regressing directly into the RGB-pixel space. While such models are shown to be useful for semantic segmentation \cite{Patraucean2016}, planning robot motion \cite{Finn2016a} or generating videos \cite{Vondrick2016}, they do not create a representation of an episode that can later be reconstructed.

\section{Method}
\subsection{The Neural Network Model}
\label{method:networkmodel}

\begin{figure*}[h]
       \centering
       \vspace{4pt}
       \includegraphics[width=0.66\paperwidth]{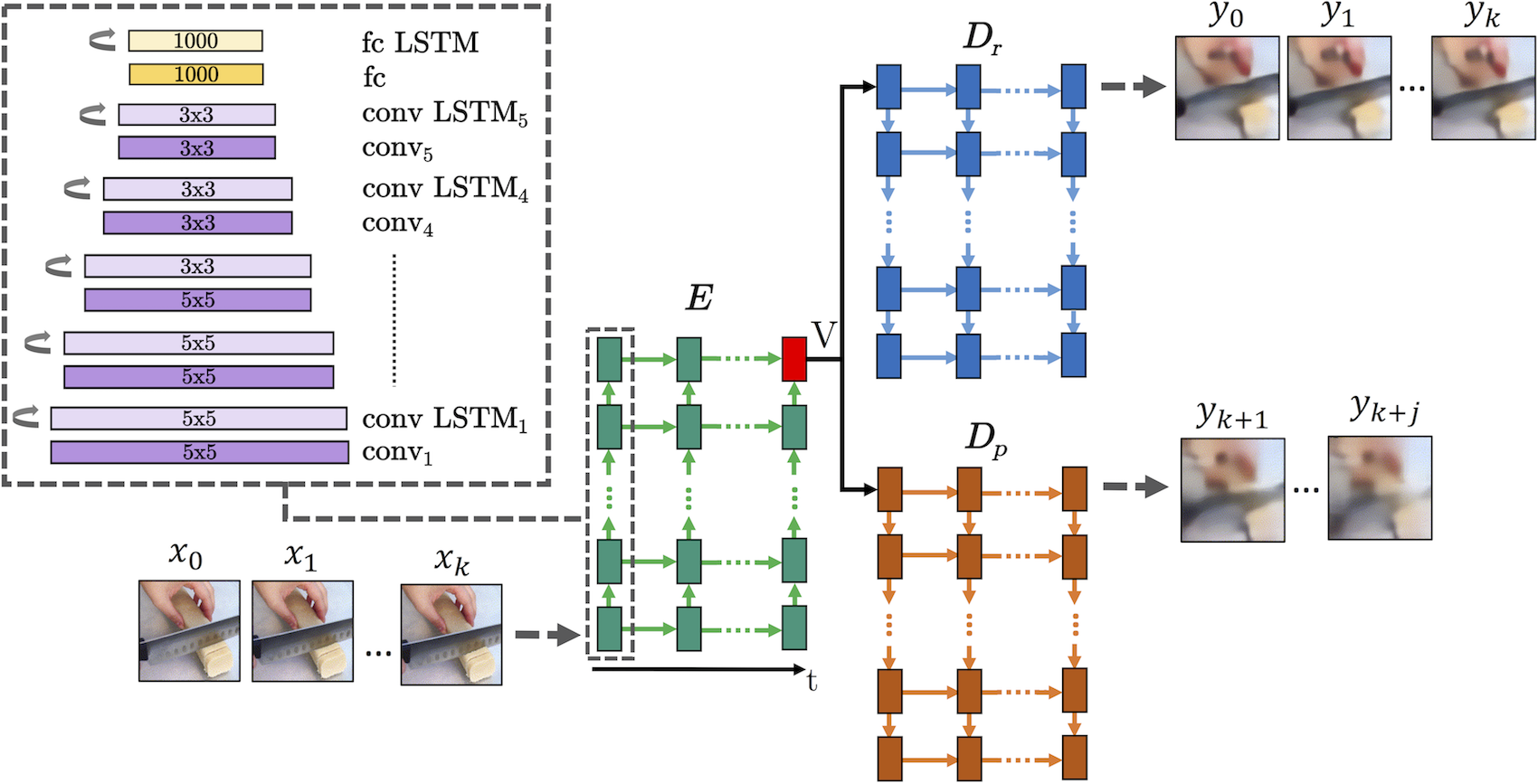}
        \caption{Structure of the proposed composite encoder-decoder network. It represents the shape of an unrolled network over multiple time steps. The encoder $\mathit{E}$ receives multiple video frames as input and maps them into a latent vector space. The resulting vector representation $\mathit{V}$ (highlighted in red) is  forwarded to the two decoder networks. The first decoder ($\mathit{D}_{r}$) is trained to reconstruct the video frames that were provided to the encoder while the second decoder ($\mathit{D}_{p}$) attempts to predict the future frames. The dashed box on the left depicts the layers of the encoder network. The label inside each layer denotes the kernel size of the convolutional layer or the number of hidden units of the fully connected (fc) layer, respectively. \vspace{-8pt}}

\label{fig:model}
\end{figure*}

In this section, we describe our neural network model and the methods applied for comparing and matching visual experiences in the latent space.
The network architecture is illustrated in Fig.~\ref{fig:model}. 

We were inspired by the composite encoder-decoder architecture introduced in \cite{Srivastava2015}. Our proposed model conceptually combines an autoencoder with future frame prediction and consists of one encoder and two decoders (see Fig.~\ref{fig:model}). Different from the model proposed by \cite{Srivastava2015}, we utilize convolutional LSTM cells to capture both spatial and temporal information. In our model, a visual experience is represented as a sequence of consecutive video frames $X=X_r \Vert  X_p= x_1,..,x_k \Vert x_{k+1},...,x_{n}$, wherein $X_r$ is the first part of the frame sequence until frame $x_k$ and $X_p$ represents the remaining $n-k$ frames.

The encoder network $E$ processes the sequence $X_r=x_1,... ,x_k$ and projects it into a latent vector space, yielding a representation of the given frame sequence as a single latent vector $V$ as:
\begin{equation}
V=E(X_r)\mathpunkt
\end{equation}

Subsequently, the vector $V$, indicated in red in Fig.~\ref{fig:model}, is forwarded to both decoders independently which receive the latent vector representation as input and construct a sequence of video frames in return. The first decoder, \ie the reconstruction-decoder $D_{r}$, attempts to recover the frames $X_r=x_1,..,x_k$ that were initially provided to the encoder. Therefore, $D_{r}$ is trained to output a frame sequence $Y_{r} = y_1, ..., y_k$ that matches $X_r$, such that
%
\begin{equation}
D_{r}(V) = Y_{r} = y_1, ..., y_k
\mathpunkt
\end{equation}
%
The second decoder, the so-called prediction-decoder $D_{p}$, attempts to predict the future frames $Y_{p} =  y_{k+1}, ..., y_n$ as, 
%
\begin{equation}
D_{p}(V) = Y_{p} = y_{k+1}, ..., y_n
\mathpunkt
\end{equation}
%
During training, $X_p$ is employed as ground truth for assessing how good the predictions $Y_{p}$ are and for also computing the error.
It is important to note that for determining the reconstruction and prediction error during training, both image sequences $X_r$ and $X_p$ are used. However, during test time, only $X_r$ is fed into the encoder network. 

The core idea of the proposed network structure rests upon the latent space vector $V$ being the only linkage between $E$ and both $D_{r}$ and $D_{p}$. The two decoder networks solely rely on $V$ as their only source of information to reconstruct a given scene and predict future frames. To obtain robust reconstructions and future frame predictions, the encoder is forced to compress the entire video frame sequence $X_r$ into a comparably low-dimensional latent representation $V$ and, at the same time, to preserve as much relevant information as possible.
$E$ and $D_r$ together constitute an autoencoder architecture, requiring that relevant information is preserved throughout the network. However, this only involves remembering the frame sequence $X_r$ but not necessarily requires to capture abstract concepts such as temporal dynamics of objects or actors. By adding the frame predictor which has to extrapolate motions into the future, the encoder must capture higher-level concepts like the scene dynamics in $X_r$ and embed abstract concepts such as trajectories in $V$ so that $D_p$ can properly infer possible future frames.

The input and output frames $x_i$ and $y_i$ used in this work have a resolution of $128\times128$ pixels and 3 color channels.

Since the main task of the network is to capture spatio-temporal concepts, we make use of convolutional LSTM cells \cite{Shi2015}. 
The encoder network $E$ is comprised of a stack of convolution LSTM and normal convolution layers (henceforth referred to as \emph{convLSTM} and \emph{conv} layers) in alternating order (see Fig.~\ref{fig:model}). While the conv layers are operated with a stride of 2 in order to reduce the spatial size of the feature maps, the convLSTM layers preserve the spatial size and forward information to the next time step through their hidden state and cell state. After the alternating series of conv and convLSTM layers in the encoder, we add a fully connected layer, followed by a fully connected LSTM layer (fc LSTM) to the stack.

With the fc LSTM cell being the top layer of $E$ and as LSTM cell connected over time, its cell state $c_i$ and hidden state $h_i$ represent the video frame sequence until the current time step $i$. Once the entire frame sequence $X_r = x_1, ..., x_k$ is processed by the encoder, the hidden state $h_k$ and cell state $c_k$ of the fc LSTM cell at time step $k$ are extracted and concatenated, yielding the latent vector $V = h_k \Vert c_k$.

Both decoders have the inverted structure of the encoder, meaning that transposed convolution layers are used to increase the spatial size of the feature maps throughout the decoding layers until the full video frame resolution of $128\times128$ is recovered. 

To compute the error $\mathit{L}$ during the network training, we use a linear combination of image reconstruction loss ($\mathit{L}_{mse}$) and gradient difference loss ($\mathit{L}_{gd}$) \cite{Mathieu2015} functions as follows  
\begin{equation}
\mathit{L}=(1- \eta)~ \mathit{L}_{mse} + \eta~ \mathit{L}_{gd}
\mathkomma
\label{eq:combinedError}
\end{equation}
where we set $\eta=0.4$ to trade off between the two loss functions
%
\begin{equation}
\mathit{L}_{mse} = \frac{1}{n}\sum_{i=1}^{n}\lVert{y_i - x_i}\lVert^2_2
\mathkomma ~~ \text{and}
\label{eq:l2}
\end{equation}
\vspace{-10px}
\begingroup
\begin{equation}
\small
\begin{split}
\mathit{L}_{gd} = \frac{1}{n} \sum_{i=1}^{n}\sum_{u,v}\lVert\lvert x_{i_{u,v}} - x_{i_{u-1,v}} \vert - \vert y_{i_{u,v}} - y_{i_{u-1,v}}\rvert\rVert^2_2 +
\\ \lVert\lvert x_{i_{u,v-1}} - x_{i_{u,v}} \rvert - \lvert y_{i_{u,v-1}} - y_{i_{u,v}}\rvert\rVert^2_2
\mathpunkt
\end{split}
\label{eq:gdl}
\end{equation}
\endgroup
The loss is computed over all ground truth frames $x_i$ in $X=X_r \Vert  X_p$ and the output frames $y_i$ in $Y=Y_r \Vert Y_p$ which are produced by $D_{r}$ and $D_{p}$. The reconstruction loss $\mathit{L}_{mse}$ compares the generated images $y_i$ and ground truth images $x_i$ in a pixel wise manner. When solely trained with $\mathit{L}_{mse}$ loss, neural network models that regress on images are prone to linear blurring and unstable to small image deformations \cite{Ranzato2014}. In contrast, the $\mathit{L}_{gd}$ loss compares the horizontal and vertical image gradients of $x_i$ and $y_i$, thereby penalizing blurriness and enforcing sharper edges \cite{Mathieu2015}.

The described neural network is trained with mini-batch gradient decent using the adaptive learning rate method ADAM \cite{Kingma2014a} in conjunction with an exponentially decaying learning rate schedule.
After each network layer except the last encoder and decoder layer (\emph{since these are the output layers}), we use layer normalization\cite{Ba2016} and dropout with a dropout rate between $10\%$ and $20\%$. In order to force the encoder to use the entire latent vector space and produce distinct representations $V$, we add Gaussian noise $N(0,\sigma)$ with $\sigma=0.1$ to the latent vector $V$ during the training, before forwarding $V$ to the decoder networks. In all of our experiments, the vector $V$ has a dimension of $2000$. The specifications of the neural network model have been determined with heuristic hyper-parameter search using train/test-splits of the original training data of the two datasets.
The source code and experimental data are publicly available on the supplementary web page{\footnote{h2t-projects.webarchiv.kit.edu/projects/episodicmemory \label{website}}}.

\subsection{Matching Visual Experiences in the Latent Space}
\label{method:matching}

One of the central contributions of our work is to compare visual experiences based on their conceptual similarities encoded in the latent space. Given a new visual experience, we can retrieve the most similar episodes from the episodic memory that holds the hidden representations $V_i$ of episodes experienced in the past.
We use the cosine similarity to measure the similarity of latent vectors. To find the best matches in the latent space, we compute the cosine similarity between the query representation $V_q$ and each of the $V_i$ in the memory. Finally, the $n$ memory instances corresponding to the $V_i$ with the highest cosine similarity are retrieved from the memory.


\section{Experimental Evaluation}

We evaluate the hypothesis that the encoder-decoder network creates latent representations that embed the inherent dynamics and concepts of a provided visual episode. 

For this purpose, we train the neural network in an unsupervised fashion on two video datasets and analyze the similarity structure within the latent space. We assess the model's abilities to reconstruct the past episode from the latent representation and predict future frames. Subsequently, we benchmark the proposed matching / retrieval mechanism against other state-of-the art approaches and test its robustness in a robotic application.


\subsection{Datasets}

For the evaluation of our methods we use the large-scale labeled video datasets ActivityNet \cite{Heilbron2015} and 20BN-something-something (from now on referred to as 20BN) \cite{Goyal2017}.
We favored these two datasets over other popular datasets like UCF-101 \cite{UCF101} and HMDB-51 \cite{Kuehne2011} since our emphasis is reasoning, planning and executing of robotic tasks in indoor household environments of which the latter two datasets contain fewer relevant scenes.

The ActivityNet dataset \cite{Heilbron2015}, a benchmarking corpus for human activity understanding, consists of $10,024$ training and $4926$ validation video snippets collected from YouTube. It is organized in 93 higher level categories that comprise $203$ different activity classes involving activities such as household work, sports and personal care.

While ActivityNet targets higher-level concepts like ``vacuuming the floor" and ``shoveling snow" that embed semantic meaning, the 20BN dataset focuses on detailed physical properties of actions and scenes. It contains $174$ classes such as ``Pushing something from left to right" and ``Putting something into something". The core challenge of this novel dataset is that the type of involved objects as well as the background setting of a given scene only play a neglectable role. Rather than recognizing familiar items and scene backgrounds, the neural network needs to understand the physical composition and motion within the video clips. The dataset consists of $86,017$ training and $11,522$ validation videos in total.

\subsection{Training}

We train the neural network on the respective training split of the ActivityNet (actNet) and 20BN dataset. Both models are trained with $n=10$ frames per video (selected equally spaced). The first $k=5$ frames are fed into the encoder $\mathit{E}$ to be reconstructed by $\mathit{D}_r$ while the last five frames are employed as ground truth for the future frame predictor $\mathit{D}_p$. 

The choice to only use 10 frames, sampled equally spaced, during training results from hardware constraints (i.e. GPU RAM) while under these constraints attempting to maximize the video length that can be captured.


\subsection{Conceptual Similarity and Proximity in the Latent Space}
\label{sim_matr}
To examine our hypothesis that conceptually similar videos are mapped into the same region of the latent space, we compute the pairwise cosine similarities (see section~\ref{method:matching}) of latent vectors. To measure the conceptual similarity, we use the class labels provided in the datasets as the proxy value and assume that videos belonging to the same class are conceptually similar. We generate the latent representation for each video in the validation split of both ActivityNet and 20BN datasets and subsequently compute all pairwise cosine similarities between the latent vectors.

\begin{figure}
\captionsetup[subfigure]{justification=justified, singlelinecheck=false}
\centering
	\vspace{5pt}
    \begin{subfigure}[h]{0.2\textwidth}            
            \includegraphics[width=0.95\linewidth]{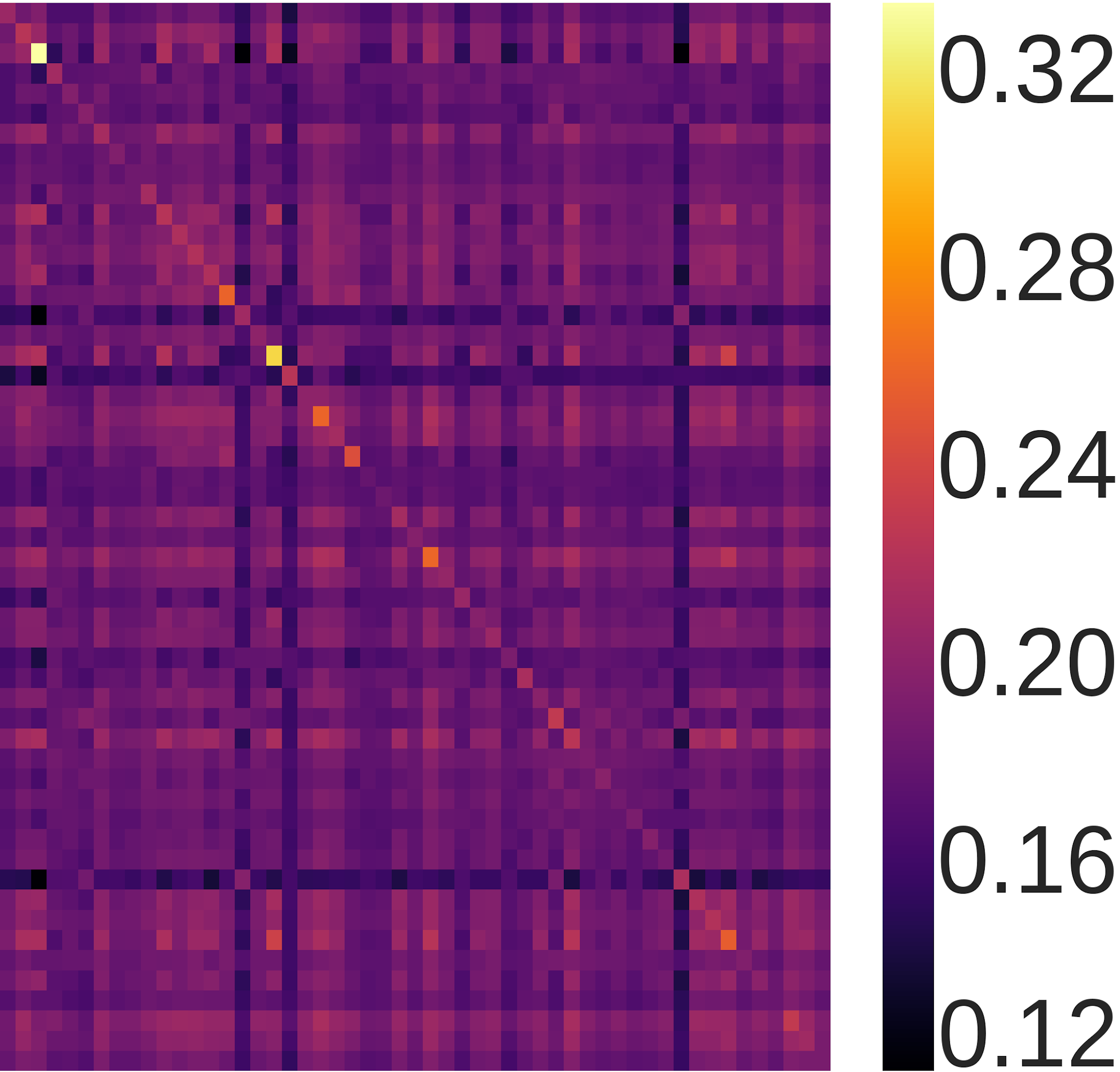}
        
            	\caption{latent vector $V$}

            \label{fig:sim_matrices_1a}
    \end{subfigure}%
    \begin{subfigure}[h]{0.2\textwidth}
            \includegraphics[width=0.95\linewidth]{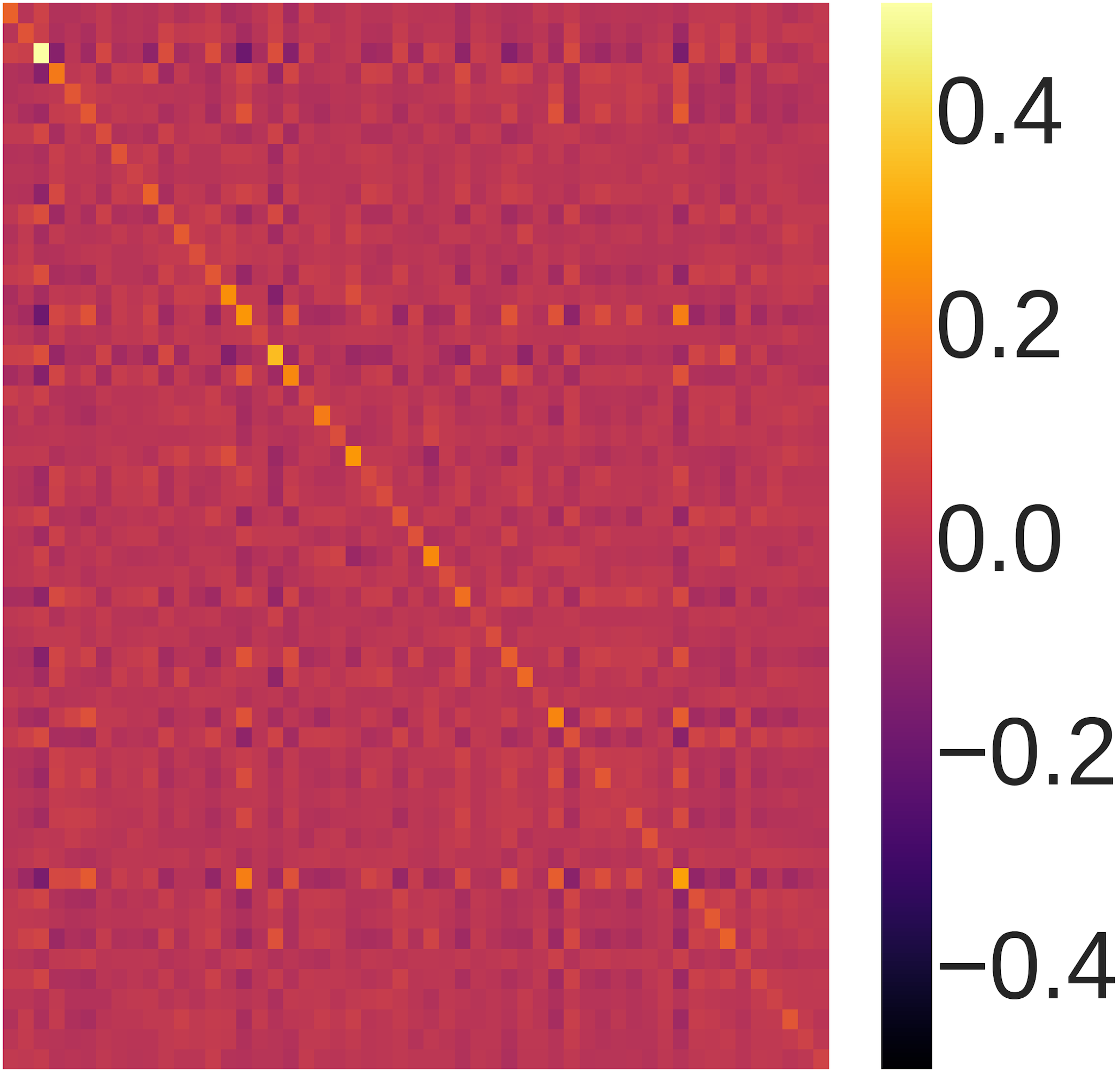}
            \caption{200 PCA components}
            \label{fig:sim_matrices_1b}
    \end{subfigure}
    \begin{subfigure}[b]{0.2\textwidth}            
            \includegraphics[width=0.95\linewidth]{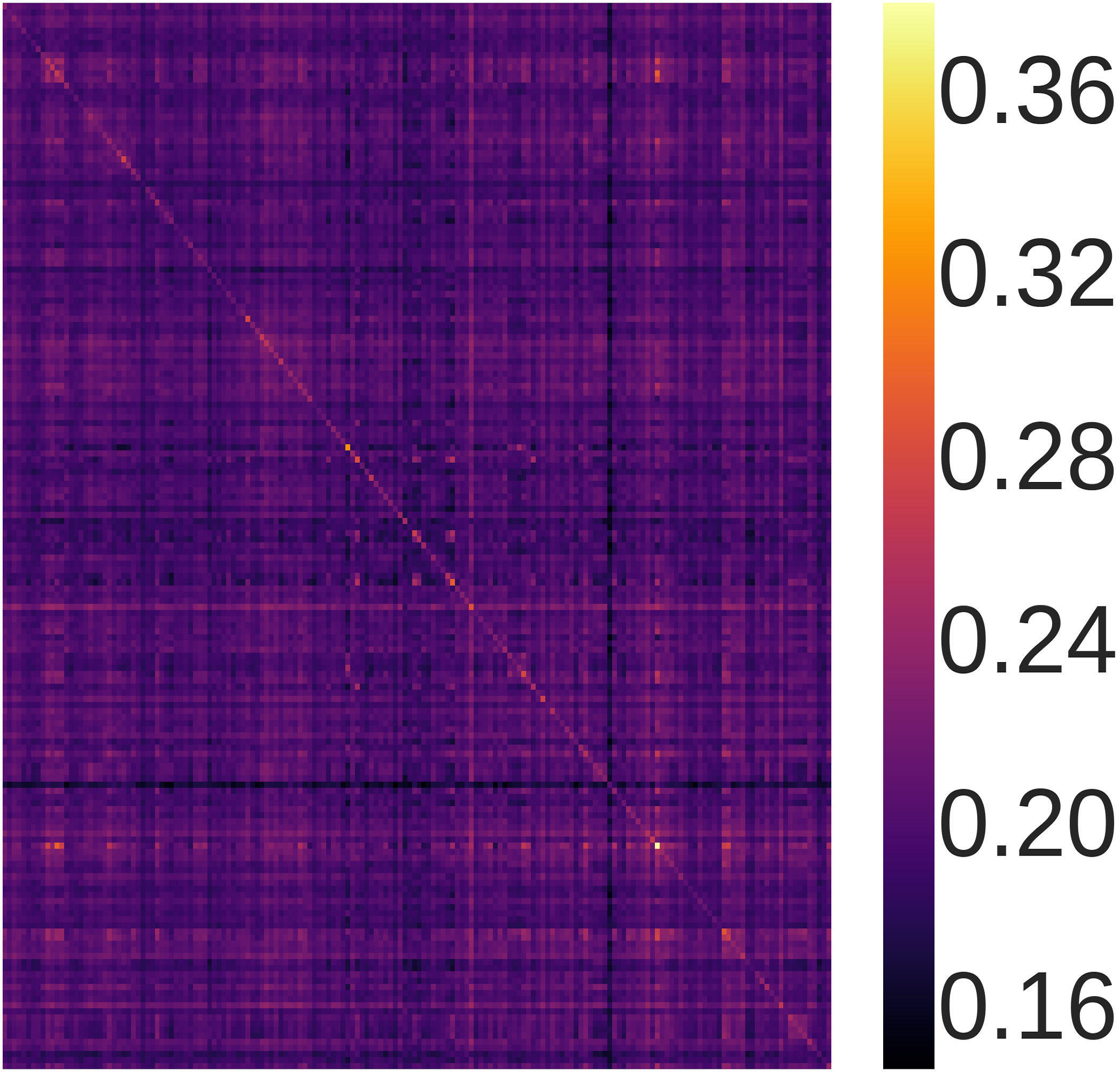}
            \caption{latent vector $V$}
            \label{fig:sim_matrices_2a}
    \end{subfigure}%
    \begin{subfigure}[b]{0.2\textwidth}
            \centering
           \includegraphics[width=0.95\linewidth]{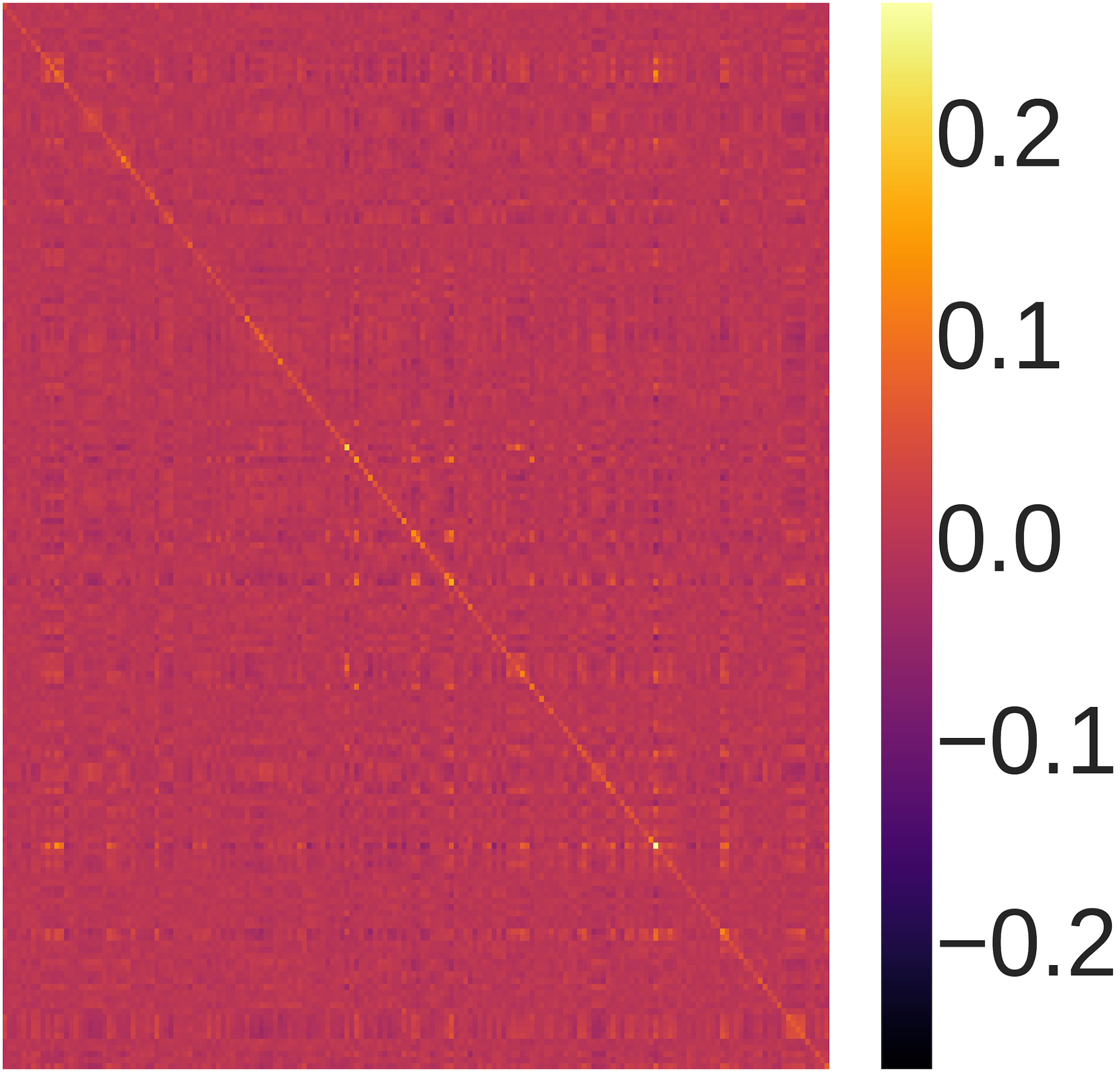}     
            \caption{200 PCA components}
            \label{fig:sim_matrices_2b}
    \end{subfigure}    
    
    \caption{Cosine similarity matrices for ActivityNet (top) ($93\times 93$) and 20BN (bottom) ($174\times 174$). PCA (right column) yields a more favorable distribution in the latent space. The figures are available in full resolution and with class labels on our website$^{\ref{website}}$.}
\label{fig:sim_matrices}
\end{figure}

In Fig.~\ref{fig:sim_matrices}, the results are visualized as similarity matrices where each row and column of a matrix corresponds to a class label and each entry represents the mean pairwise cosine similarity between the latent vectors belonging to the respective classes.
The class labels are arranged horizontally and vertically in the same order, ensuring that the diagonal elements of the matrix depict intra-class similarities and off-diagonals represent inter-class similarities. Fig.~\ref{fig:sim_matrices} shows that in each matrix the intra-class similarity (diagonal elements) is considerably higher than the inter-class similarity. This is a clear indication that conceptual similarity of videos is reflected by the proximity of their vector representations in the latent space. Consequently, our proposed model 
captures high-level action concepts within two different datasets, although the model is trained in an \emph{unsupervised} fashion and thus has never seen any class labels.

Since the latent representation must embed all information for the decoders necessary to reconstruct and predict frames, it may also encode details (\eg colors and shapes) in the background that are irrelevant for describing actions. To compile the information embedded in the latent representation to a subset that is more relevant for optimally separating the different classes within the latent space, we apply principal component analysis (PCA) on the mean latent vectors of each class. We assume that less important features are identically distributed in all the classes and thus share approximately the same mean value when averaged over the class. Hence, transforming the latent space towards the principal components computed on the covariance matrix of the mean vectors emphasizes relevant features and neutralizes less important features.
Fig.~\ref{fig:sim_matrices_1b} and \ref{fig:sim_matrices_2b} show that transforming the latent representations with PCA leads to a better distribution of latent vectors, pushing conceptually similar representations closer together while keeping representations of different classes farther apart.

\begin{figure*}[!t]
\centering
\vspace{3pt}
\includegraphics[width=0.7\linewidth]{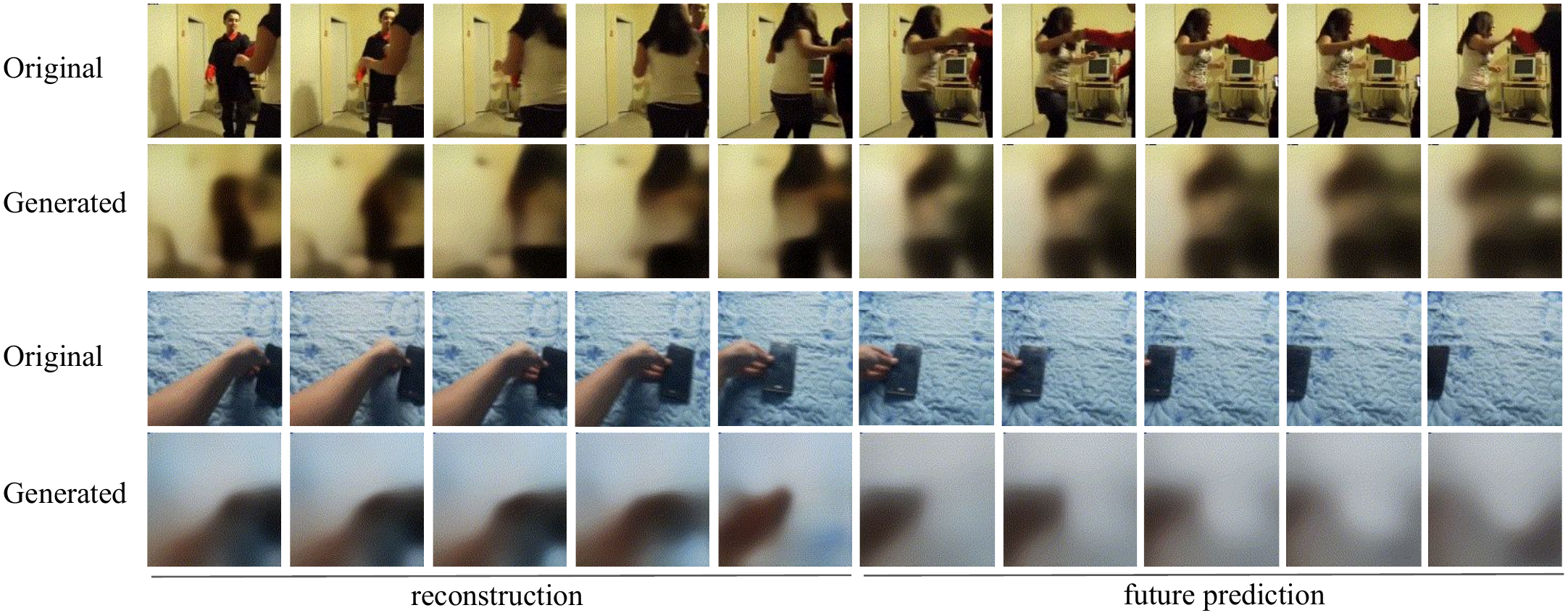}
  \caption{Frame reconstruction and future frame prediction of our model compared to the original frame sequence. Top: Validation sample from ActivityNet. Bottom: Validation sample (pulling sth. from right to left) from 20BN.
\vspace{-10pt}}\label{fig:reconst_pred}
\end{figure*}

\subsection{Frame Reconstruction and Future Frame Prediction}

\begin{figure}[!h]
\centering
\addedimage{
\includegraphics[width=0.80\linewidth]{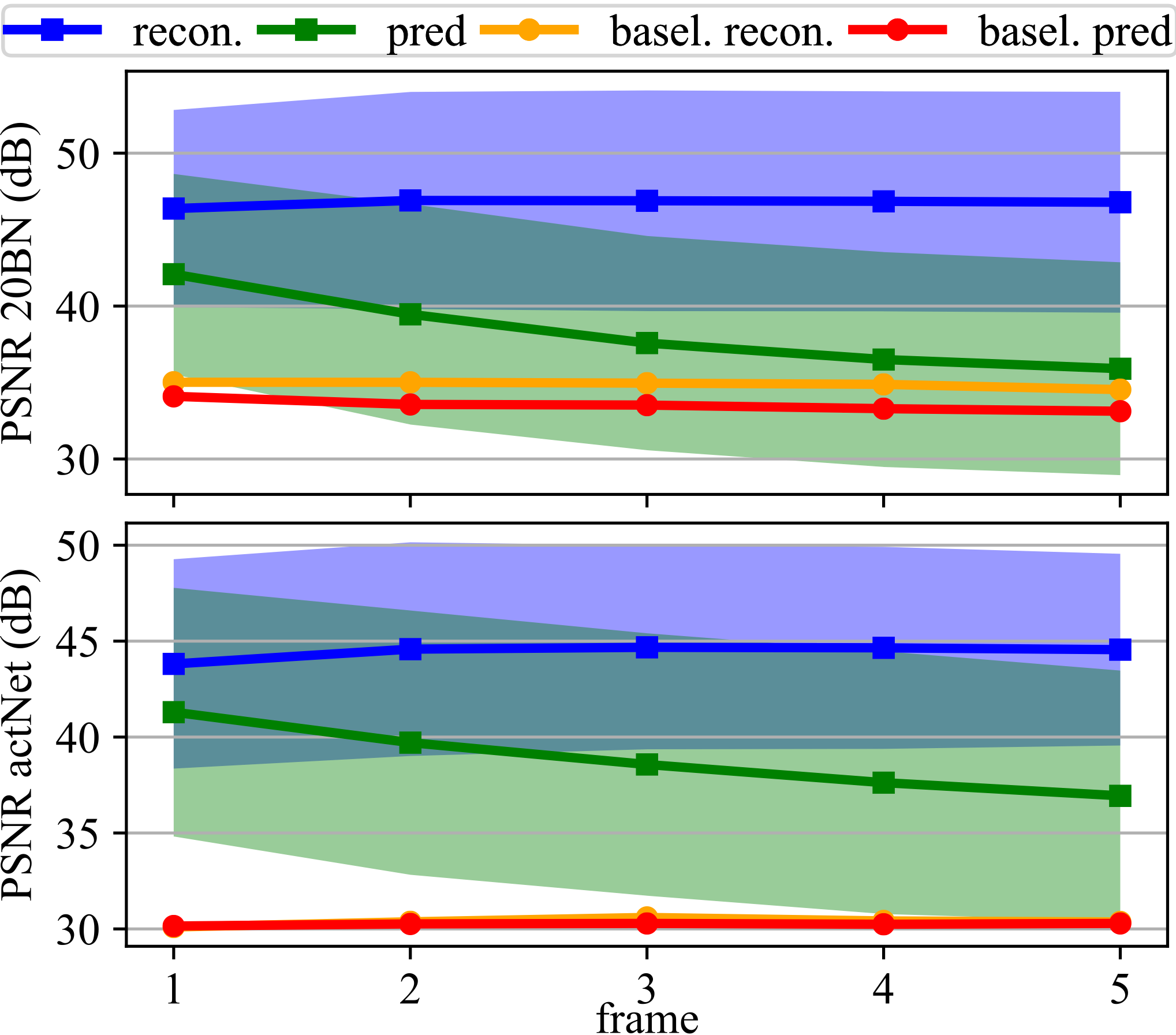}
}
  \caption{Peak-Signal-to-Noise-Ratio (PSNR) for each of the 5 reconstructed (blue) and predicted (green) frames generated by the 20BN and actNet model averaged over the entire validation dataset. The shaded regions depict the corresponding standard deviation. Additionally using the channel-wise mean of input frames as reconstruction/prediction is reported as simple baseline (red/orange).
\vspace{-10pt}}\label{fig:psnr_chart}

\end{figure}

By reconstructing video frames and predicting upcoming frames, the network resembles episodic memory-like capabilities.
Fig.~\ref{fig:reconst_pred} depicts generated video frame sequences from both ActivityNet and 20BN for a qualitative assessment.
To evaluate the quality of the reconstructed and predicted frames, we compute the Peak-Signal-to-Noise-Ratio (PSNR) between the original frames $X$ and generated frames $Y$ as proposed in \cite{Mathieu2015}. Fig.~\ref{fig:psnr_chart} depicts the PSNR for each of the 5 reconstructed and predicted frames, averaged frame-wise over the entire respective validation samples of both datasets. We additionally compare our results to a simple baseline by computing the PSNR between the ground truth frames and the mean of the input frames. Results indicate that the reconstruction quality is significantly higher than the quality of the predicted frames. Also, the PSNR is roughly constant throughout the reconstructed frames whereas for the predicted future frames it decreases over time. 
The expected decline in prediction quality is very much due to the increase in the uncertainty about the future over successive time steps. Moreover, the frame prediction results in Fig. \ref{fig:reconst_pred} indicate that the decline in PSNR is mainly caused by the increase in blurriness while e.g. the object motions are still extrapolated correctly into the future. Hence, this successive increase in blurriness predominantly constitutes a general limitation of regression into the pixel space and gives little to no indication about the  semantic richness of the latent representations.

\subsection{Matching and Retrieving Visual Episodes}

To investigate the matching and retrieval of visual episodes introduced in section~\ref{method:matching}, we compare our approach to standard baselines as well as state-of-the-art action descriptors. The benchmarking comprises Fisher vector encodings of CNN, SIFT and STIP features as well as LSTM encodings to represent and match visual experiences.

As proposed in \cite{Doshi2015}, we compute Fisher Vectors based on GMMs in order to create a visual vocabulary from CNN features of the video frames. In particular, we select 5 equidistant frames from the videos and compute VGG-fc1 \cite{VGG} as well as ResNet-50 \cite{Resnet} features.
Furthermore, we compare our approach against the composite LSTM network introduced in \cite{Srivastava2015}. We train their LSTM network using VGG-FC1 and ResNet-50 features. In all our experiments we used the default parameters coming with the publicly available source codes.

To quantitatively benchmark our approach against the baselines, we phrase the matching and retrieval of memorized episodes as a document retrieval problem.
Thereby, we assume that a retrieved episode is only relevant if it originates from the same action category as the query episode. For evaluating the performance of retrieving relevant visual episodes, we report the precision of the first match and the mean average precision (mAP). Since the setting is purely unsupervised, the precisions reported in this context are not to be confused with precisions for a classification task. Overall, we use the ActivityNet and the 20BN dataset to train and evaluate the different methods. For computing the performance metrics, we split the validation set of the respective dataset into 5 shuffled folds. In each fold, 80\% of the videos are used as memory whereas the remaining 20\% are used to query the memory. The results illustrated in Table \ref{table:baslines} are averaged over the 5 folds.

The evaluations were conducted with various numbers of GMM components and two different distance metrics (i.e. Euclidean and cosine) for matching the video encodings. We want to emphasize that we always report the best results out of these experimental evaluations.

As the results in Table \ref{table:baslines} indicate, representations of conventional descriptors such as SIFT and STIP seemingly lack the representational capacity to capture abstract spatio-temporal concepts in videos such as action. In contrast, using CNN features yields significantly higher average matching precision. Our proposed method has the highest precision values and notably outperforms state of the art approaches with substantially deeper networks such as ResNet-50.
For completeness, Table \ref{table:baslines} also reports the matching precisions with 200 PCA components. Similar to the observations discussed in section \ref{sim_matr}, PCA further improves the matching. The precisions on ActivityNet are substantially higher since it has a) fewer categories than the 20BN dataset and b) richer features among the categories.

\begin{figure*}[!t]
\centering
\vspace{2pt}
\includegraphics[width=0.86\linewidth]{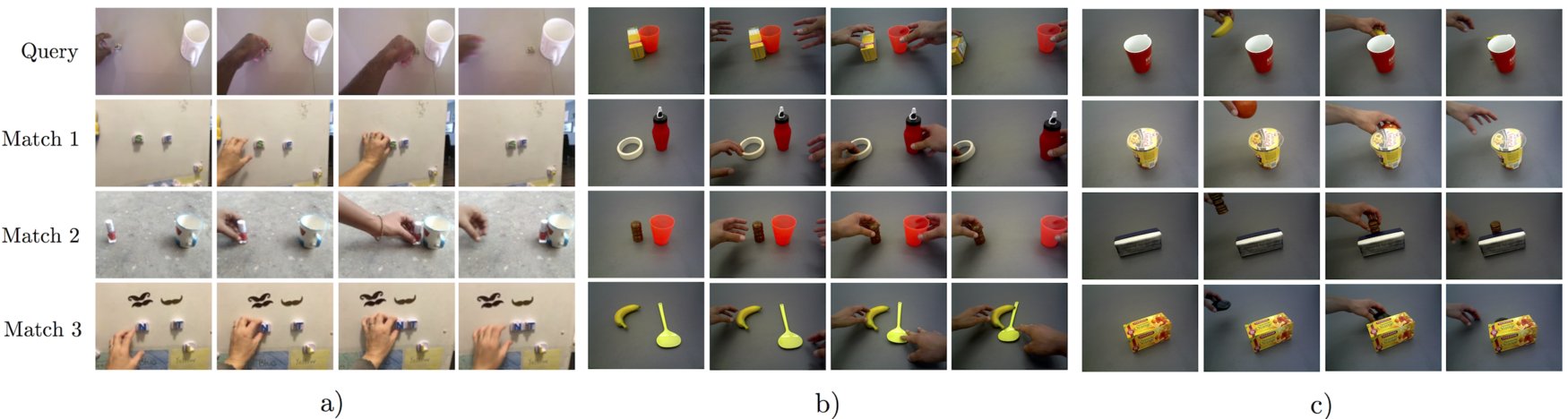}
  \caption{The proposed matching and retrieval mechanism evaluated on different visual episodes of object manipulations. The figure depicts three exemplary query episodes and the corresponding 3 closest matches in the latent space of our proposed network. The latent representations are generated by the 20BN model. While (a) shows a query on the 20BN dataset, (b) and (c) comprise human demonstrations for the \armarIIIa robot (see section \ref{sec:case_study} ). 
\vspace{-10pt}}\label{fig:matching_comparison}

\end{figure*}

In addition to the quantitative benchmark, we qualitatively examined the matching and retrieval results. The matching appears to be conceptually consistent, predominantly yielding visual episodes with closely related action types and settings. Fig~\ref{fig:matching_comparison}~(a) shows sample matching results of our approach on the 20BN dataset. In most cases of categorical mismatches, the retrieved episode is closely related to the action type of the query episode, e.g. "pushing sth off of sth" and "moving sth away of sth" or "lifting sth up" and "moving sth away from sth". However, we also observe that the background color biases the matching results, meaning that videos with dark background tend to be matched to videos in memory of the same kind. The same applies to bright videos respectively.

All in all, our proposed model achieves the best matching results among the compared approaches. The qualitative study shows that, beyond the strict categorical setting imposed in the baseline comparison, our mechanism predominantly produces consistent matches. It is also important to note that our approach is the only one that can reconstruct the visual episode given the encoding.

\begin{table}
\centering
\caption{Benchmark results showing the matching and retrieving performance of our approach against the baselines. We report the precision for the first match and the mean average precision (mAP) for retrieving the 3 closest matches. Fisher Vectors are abbreviated to FV.}
\label{table:baslines}
\begin{tabular}{@{}lcc|cc@{}}
\toprule
 & \multicolumn{2}{c|}{\textbf{ActivityNet (in \%)}} & \multicolumn{2}{|c}{\textbf{20BN-sth.-sth. (in \%)}} \\ \midrule

\textbf{Model and Features} & \multicolumn{1}{c}{\textbf{Precision}} & \multicolumn{1}{c}{\textbf{mAP}} & \multicolumn{1}{|c}{\textbf{Precision}} & \multicolumn{1}{c}{\textbf{mAP}}  \\ \midrule

SIFT FV & 5.27 & 3.76  & 1.05 & 0.64  \\

STIP FV \cite{STIP} & 5.47 & 3.71  & 3.37 & 2.67 \\

ResNet-50 FV \cite{Doshi2015}  &  32.31 & 23.23 &  6.08 & 3.98  \\

VGG-16 FV \cite{Doshi2015}  & 24.30 & 18.19 & 5.56 & 3.62 \\

ResNet-50 LSTM \cite{Srivastava2015} & 36.29 & 26.32 & 10.08 & 7.20  \\

VGG-16 LSTM \cite{Srivastava2015} & 17.18  & 12.16 & 4.51 & 2.82 \\

\textbf{ours (no PCA)}  & \textbf{44.31} & \textbf{26.93} & \textbf{11.63}  & \textbf{8.12} \\
\midrule
ours with PCA (200) & 45.55 & 28.18 & 11.81 & 8.32 \\
 \bottomrule
\end{tabular}
\end{table}

\subsection{Robot Manipulation Learned From Episodic Memory}
\label{sec:case_study}

Results in the previous section indicate that given a query action, our proposed network provides a robot with the ability to remember similar scenarios from a large video corpora. In the following, we show how our episodic memory approach can facilitate case-based reasoning for robot manipulation in unstructured real world environments.
 
For this purpose, we record 120 visual episodes in which a human subject is demonstrating to our humanoid robot \armarIIIa \cite{Asfour2006} how to perform 10 different manipulation actions such as ``pushing two objects closer to each other" and ``putting something behind something" (see Fig.~\ref{fig:matching_comparison}~(b-c)).
The reason of introducing this new dataset is twofold: First, we attempt to evaluate the scalability of our approach to new datasets that have much less training data. Second, for the purpose of action execution we require the depth cue which is missing in the ActivityNet and 20BN datasets.
We store 100 of our new visual episodes to form the memory, whereas the remaining 20 episodes are introduced as queries to test the matching and retrieval mechanism with \armarIIIa. The visual episodes are fed through the encoder of the trained 20BN model, thereby receiving its respective latent representations. The cosine similarity in the latent space is then computed based on the first 50 principal components of the latent representations (see section \ref{method:matching}). Fig.~\ref{fig:matching_comparison}~(b-c) illustrate two exemplary query episodes and the corresponding 3 closest matches in the latent space from our recordings.

So far, the matching and retrieval mechanism is evaluated on manipulation action videos where spatio-temporal cues are implicitly embedded. We further investigate whether static scene frames can trigger recalling of past episodes, which is an even harder problem since effectively only one frame is given as a cue. This gives a high chance to robots to autonomously predict and even execute an action that can possibly be performed in the observed scene. \added{Matching results for static scene queries can be viewed on our supplementary website$^{\ref{website}}$}
We conduct a pilot study to explore the use of our method. 
 
\begin{figure}[!b]
\centering
\includegraphics[width=1.0\linewidth]{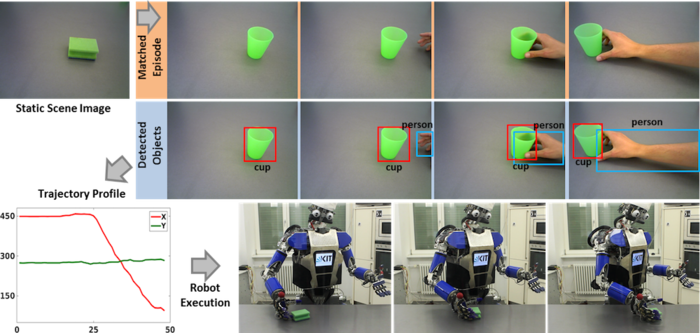}
  \caption{ Robot execution of a matched pushing action. }
  \label{fig:robot_execution}
\end{figure}

Fig.~\ref{fig:robot_execution} illustrates a scenario where the robot \armarIIIa is observing a scene with a sponge on the table. 
Acquired images of this static scene are sent to our matching and retrieval mechanism which returns the matched episode where a subject is demonstrating ``pushing a green cup". 
Next, we apply a real-time object detector \cite{YOLO} to detect and track all objects in the recalled episode. 
This process yields the extracted pushing trajectory that the subject is following.
The tracked motion is then learned by dynamic movement primitives \cite{Ijspeert2002} to be further processed by the robot in order to execute the same pushing motion on the perceived sponge. 
Fig.~\ref{fig:robot_execution} depicts sample frames from the best matched episode and detected objects together with the computed motion profile and snapshots from the robot execution of the recalled pushing action.
See the supplementary movie showing the entire robot execution.

This experiment clearly supports our hypothesis that the proposed network model can help robots autonomously trace previous observations and select the one that matches best to the currently observed scene even without necessarily requiring any temporal cue. The robot can transfer relevant knowledge, \eg the motion profile, from previous experiences to further apply the remembered action to novel objects in the scene.
Learning comparable vision triggered behaviors through reinforcement learning would require thousands of trials, while such a case-based reasoning approach only requires one memory-instance of a similar action. Hence, these findings play a vital role in cognitive robotics to infer possible actions, reason about the action consequences, and even generate actions by transferring knowledge from the past experiences in a data efficient manner.

  
\vspace{-1pt}
\section{Conclusion}
\vspace{-1pt}
We proposed a deep neural network implementing an episodic memory. Given a set of training data, the proposed network first generates subsymbolic representation of action episodes. Such a latent encoding can be used to  distinguish actions, reconstruct memorized episodes, and predict future frames based on the spatio-temporal features extracted by the deep architecture. 
We show that conceptual similarity of videos is reflected by the proximity of their vector representation in the latent space.
Using this property of the latent space, we introduce a matching and retrieval mechanism, which enables the recollection of previously experienced visual episodes. Benchmarking our proposed mechanism against a variety of action descriptors, we show that our model outperforms other state-of-the-art approaches in terms of matching precision. 
We conduct various experiments showing that the proposed framework can help extending the cognitive abilities of a humanoid robot such as action encoding, storing, memorizing, and predicting.

Since the memory instances in our approach are purely visual, the adaptation of stored actions on a robot to new scenario is not straightforward. 
\added{For instance, our robot manipulation pilot study requires an auxiliary object detector for trajectory extraction.} In future work it would be desirable to embed relevant action information such as trajectories into the latent representation and directly use them for action adaptation and execution. Another limitation of our approach is the observed background bias. To make the model less sensitive to different backgrounds it might be promising to avoid pixel-level predictions and instead employ segmentation masks as reconstruction targets.

To the best of our knowledge, this is the first comprehensive study that attempts to encode visual experiences not only for matching and retrieving purposes but also for prediction and reconstruction in a longer time scale. Comparable work such as \cite{Doshi2015, Srivastava2015} can only achieve unsupervised action matching without reconstructing the memorized video episodes. Video prediction models, introduced in \cite{Finn2016a, Patraucean2016}, can only predict a single future frame at a time. Thus, the mentioned approaches lack key features to resemble an episodic memory. Our model overcomes the architectural drawback of \cite{Srivastava2015} and comprises the full episodic-memory capabilities described above. Also, we are not aware of any previous work that extensively applies an episodic memory-like framework to such large and complex datasets.

%



 


\bibliographystyle{IEEEtran}
 \bibliography{library}

\end{document}